\documentclass[11pt,a4paper]{article}
\usepackage{iclr2021_conference,times}

\usepackage{amsmath,amsfonts,bm}

\def\eqref#1{equation~\ref{#1}}

\def\1{\bm{1}}

\DeclareMathAlphabet{\mathsfit}{\encodingdefault}{\sfdefault}{m}{sl}
\SetMathAlphabet{\mathsfit}{bold}{\encodingdefault}{\sfdefault}{bx}{n}

\usepackage[hidelinks]{hyperref}
\usepackage{url}
\usepackage{xspace}
\usepackage{latexsym}

\usepackage{microtype}

\usepackage[utf8]{inputenc} 
\usepackage[T1]{fontenc}    
\usepackage{booktabs}       
\usepackage{amsfonts}       
\usepackage{amsmath}
\usepackage{nicefrac}       
\usepackage{color}
\usepackage{tikz}
\usepackage{pgfplots}
\usepackage{subcaption}
\usepackage{enumitem}
\usepackage{xparse}
\usepackage{tikz}
\usepackage{csquotes}
\usetikzlibrary{arrows.meta,arrows}
\pgfplotsset{compat=1.14}

\definecolor{g-blue}{HTML}{2E86C1}
\definecolor{g-red}{HTML}{B03A2E}
\definecolor{g-purple}{HTML}{AF7AC5}

\newcommand{\red}[1]{\textcolor{red}{#1}}
\newcommand{\ignore}[1]{}

\usepackage{array}
\newcolumntype{H}{>{\setbox0=\hbox\bgroup}c<{\egroup}@{}}

\usepackage{multirow}

\NewDocumentCommand{\gin}{ mO{} }{\textcolor{orange}{\textsuperscript{\textit{Gin}}\textsf{\textbf{\small[#1]}}}}

\newcommand{\decimal}[0]{\textsc{decimal}\xspace}
\newcommand{\character}[0]{\textsc{character}\xspace}
\newcommand{\fixedCharacter}[0]{\textsc{fixed-character}\xspace}
\newcommand{\underscore}[0]{\textsc{underscore}\xspace}
\newcommand{\words}[0]{\textsc{words}\xspace}
\newcommand{\tenBased}[0]{\textsc{10-based}\xspace}
\newcommand{\tenEBased}[0]{\textsc{10e-based}\xspace}

\title{Investigating the Limitations of Transformers with Simple Arithmetic Tasks}

\author{Rodrigo Nogueira, Zhiying Jiang \& Jimmy Lin \\
David R. Cheriton School of Computer Science\\
University of Waterloo
}

\iclrfinalcopy 

\begin{document}

\maketitle

\begin{abstract}
The ability to perform arithmetic tasks is a remarkable trait of human intelligence and might form a critical component of more complex reasoning tasks.
In this work, we investigate if the surface form of a number has any influence on how sequence-to-sequence language models learn simple arithmetic tasks such as addition and subtraction across a wide range of values.
We find that how a number is represented in its surface form has a strong influence on the model's accuracy.
In particular, the model fails to learn addition of five-digit numbers when using subwords (e.g., ``32''), and it struggles to learn with character-level representations (e.g., ``3 2'').
By introducing position tokens (e.g., ``3 10e1 2''), the model learns to accurately add and subtract numbers up to 60 digits.
We conclude that modern pretrained language models can easily learn arithmetic from very few examples, as long as we use the proper surface representation.
This result bolsters evidence that subword tokenizers and positional encodings are components in current transformer designs that might need improvement.
Moreover, we show that regardless of the number of parameters and training examples, models cannot seem to learn addition rules that are independent of the length of the numbers seen during training.
Code to reproduce our experiments is available at \url{https://github.com/castorini/transformers-arithmetic}
\end{abstract}

\section{Introduction}

Abstraction and composition are two important themes in the study of human languages, made possible by different linguistic representations. 
Although treatments in different linguistic traditions vary, representations at the lexical, syntactic, and semantic levels are a common feature in nearly all theoretical studies of human language, and until relatively recently, these representations are explicitly ``materialized'' in language processing pipelines (for example, semantic role labeling takes as input a syntactic parse).

However, with the advent of pretrained transformer models, these intermediate representations no longer have any explicit ``reality'':\ while various studies have found evidence of syntactic and semantic knowledge in these models~\citep{tenney-etal-2019-bert}, it is no longer possible to isolate, for example, a subject--verb relation in a specific part of the model.
With transformers, the {\it only} input to the model is the surface form of text combined with supplemental embeddings (e.g., positional embeddings, and in the case of BERT, segment embeddings).

What are the consequences of this exclusive focus on the surface form of text?
Some might say, nothing, as bigger models, better pretraining objectives, etc.~will lead us to models that are capable of reasoning~\citep{gpt3}.
We believe this to be an untenable position and present a case study in simple arithmetic tasks where having the right representation is the difference between a nearly-impossible-to-learn task and an easy-to-learn task.
Our work shows that it is possible to ``inject'' representations into transformer models by simple manipulations of the input sequence (in our case, explicitly enumerating the semantics of digit positions), and that doing so makes it possible for off-the-shelf models to easily perform simple arithmetic, whereas it is nearly impossible otherwise.

While we present only a case study, our findings have broader implications for various language analysis tasks:
First, although end-to-end training enabled by neural networks is a powerful tool, having the right representation is crucial also.
Second, we demonstrate a simple way in which representations can be ``injected'' into transformer models in a completely transparent manner, without any need to re-pretrain.
This work points out a path that might allow us to combine the best of both worlds:\ leveraging the power of pretraining, with additional guidance from our understanding of the problem domain.

However, we find that even explicit semantic representations have their limits. Despite our best efforts, we find that models cannot extrapolate, i.e., they fail to perform simple arithmetic when evaluated on inputs whose length distribution differs from the one seen during training.
This appears to be a problem that neither larger models, more compute, nor more data can solve.

There are, of course, many previous papers that investigate the representation of numbers and various numeric reasoning tasks in the literature.
We present related work in Appendix~\ref{section:related_work}.

\section{Methodology}
\label{section:methodology}

Our tasks are the addition and subtraction of two numbers.
We cast them as sequence-to-sequence tasks in which both inputs to the models and target outputs are treated as sequences of tokens.
For the addition task, an example input is ``What is 52 plus 148?'' and the target output is ``200''.
For the subtraction task, an example input is ``What is 20 minus 185?'' and the target output is ``-165''.

We programmatically generate training, development, and test sets of different sizes depending on the experiment.
The input template is always ``What is [number1] [operation] [number2]?'', where [number1] and [number2] are numbers randomly sampled and [operation] is either ``plus'' or ``minus''.
Below, we discuss different ways of representing [number1] and [number2] and their corresponding answer.
We use two different methods to sample numbers for training, development, and test sets, which are described below.

{\bf Balanced sampling:} To generate training and development sets, we first set the maximum number of digits $D$ and then create each example as follows:\ We first sample $d$ from $[2, D]$ and then independently sample [number1] and [number2] from $[10^{d-1},10^{d} - 1]$.
We then compute the answer according to the operation (i.e., either addition or subtraction).
This method ensures that the set will have a roughly equal proportion of $d$-digit numbers, where $d \in [2, D]$.

{\bf Random sampling:} To generate test sets, we sample [number1] and [number2] independently from $[0, 10^D - 1]$.
This results in approximately 90\% of the numbers having $D$-digits, 9\% having $(D-1)$-digits, and so on.
This unbalanced set aims at evaluating models on the largest numbers it was trained on.
We study how different sampling methods influence model effectiveness in Appendix~\ref{section:out_of_distribution}.

{\bf Metric:} Our metric is accuracy.
That is, the model receives a score of one if its output matches the target output exactly.
Otherwise, it receives a score of zero.

Our experiments use T5~\citep{2020t5}, a pretrained sequence-to-sequence model where every natural language processing task---for example, machine translation, question answering, and classification---is formulated as feeding the model some input sequence and training it to generate some output sequence.
We follow this same approach and feed the addition or subtraction question (described above) as a sequence of tokens to the model and train it to generate the answer, token by token.
We use greedy decoding as beam search showed similar effectiveness but is slower.

We train the models using the AdamW optimizer~\citep{loshchilov2018decoupled}, batches of 128 examples, and a learning rate of $0.0003$.
We experimented with all T5 model sizes except for T5-11B due to its computational cost.
We refer to T5-small, T5-base, and T5-large as T5-60M, T5-220M, and T5-770M, respectively, to easily distinguish models by their numbers of parameters.
We also experiment with ``vanilla'' (i.e., non-pretrained) transformers (see Appendix~\ref{section:position_embeddings}).

\begin{table}[t]
    \centering
    \begin{tabular}{l|r|l}
         Orthography & Example & Notes \\
         \midrule
         \decimal & 832 & default representation \\
         \character & 8 3 2 & ensures consistent tokenization \\ 
         \fixedCharacter & 0 8 3 2 & ensures consistent positions (e.g., max.\ 4 digits)\\ 
         \underscore & 8\_3\_2 & underscores provide hints on digit significance\\
         \words & eight hundred thirty-two & leverages pretraining \\
         \tenBased & 8 100 3 10 2 & easy to determine digit significance\\
         \tenEBased & 8 10e2 3 10e1 2 10e0 & more compact encoding of above\\
    \end{tabular}
    \caption{Different ways of representing numbers explored in this work.}
    \label{table:orthographies}
\end{table}

Previous studies have recognized that commonly used subword tokenization techniques today are not ideal to represent numbers~\citep{wallace2019nlp,henighan2020scaling,saxton2018analysing,lample2019deep}, although none of them studied the problem in depth.
Here, we investigate how six different number representations, illustrated in Table~\ref{table:orthographies}, impact model accuracy on the arithmetic tasks.
In our main results, we only experiment with the ``standard'' ordering of generating digits (i.e., most to least significant), but in Appendix~\ref{section:interpolation_extrapolation}, we also experimented with inverting the order.

{\bf \decimal:} Digits are represented in the Hindu–Arabic numeral form (also called decimal form).

{\bf \character:} Digits are separated by a white space, thus allowing the model to work on embeddings that always represent single digits.

{\bf \fixedCharacter:} In the character representation above, it is hard to determine the significance of a digit by relative position embeddings because relative positions change on a per example basis.
To address this, we introduce the \fixedCharacter representation in which numbers have the same maximum number of digits.

{\bf \underscore:} Digits are separated by an underscore token.
A possible advantage of this representation is that the model can learn to find the significance of a digit by counting the number of underscores to the right until the least significant digit.

{\bf \words:} Numbers are converted to words using the \textit{num2words} package.\footnote{\url{https://github.com/savoirfairelinux/num2words}}
We can anticipate two advantages in this representation:\ (1) the T5 model was pretrained on large amounts of textual data, so it likely knows that ``hundred'' is larger than ``ten''~\citep{zhang2020language}; (2) digits are surrounded by tokens that describe their significance (``hundred'', ``thousand'', etc.), thus making it easier to find which two digits in the input sequence should be added (or subtracted).

{\bf \tenBased:} Digits are separated by powers of 10, which we call position tokens.
This representation allows the model to find the significance of a digit by simply inspecting its left or right tokens.

{\bf \tenEBased:} Digits are separated by powers of 10 represented using scientific notation.
This orthography has a more compact representation for the position tokens of large numbers than the \tenBased orthography.
For example, in the \tenBased orthography, the position token of the most significant digit of a 60-digit number occupies 60 characters (i.e., ``1'' followed by 59 zeros).
In the \tenEBased orthography, this position token occupies only 5 characters (i.e., ``10e59'').

\section{Results}
\label{section:orthography_results}

\begin{figure*}[t]
\centering
\begin{tikzpicture}[scale = 0.9]
\begin{axis}[
width=0.75\textwidth,
height=0.40\textwidth,
legend cell align=left,
font=\small,
axis y line*=left,
xmin=2, xmax=30,domain=1:10,
ymin=0.0, ymax=1.0,
every axis plot/.append style={very thick},
xtick={2,5,10,15,20,25,30},
ytick={0.0, 0.2, 0.4, 0.6, 0.8, 1.0},
legend pos=outer north east,
xmajorgrids=true,
ymajorgrids=true,
xlabel style={font = \small, yshift=1ex},
xlabel=\# of digits,
ylabel= Test Accuracy,
ylabel style={font = \small, yshift=0ex}]

\addplot[
  black, mark=triangle*, red, mark options={scale=1},
  error bars/.cd, 
    y fixed,
    y dir=both, 
    y explicit
] table [x=x, y=y,y error=error, col sep=comma] {
    x,    y,       error
    2, 0.9953, 0.0027
    5, 0.9348, 0.0150
    10, 0.9571, 0.0147
    15, 0.9708, 0.0123
    20, 0.6775, 0.0349
    25, 0.5728, 0.1298
    30, 0.6288, 0.0750
};
\addlegendentry{\tenEBased ``3 10e1 2 10e0''}

\addplot[
  black, mark=square*, black, mark options={scale=1},
  error bars/.cd, 
    y fixed,
    y dir=both, 
    y explicit
] table [x=x, y=y,y error=error, col sep=comma] {
    x,    y,       error
    2, 0.9964, 0.003 
    5, 0.9462, 0.011
    10, 0.9547, 0.0073
    15, 0.9360, 0.0213
    20, 0.3222, 0.0991
    25, 0.2910, 0.1524
    30, 0.3014, 0.1088
};
\addlegendentry{\tenBased ``3 10 2''}

\addplot[
  solid, mark=triangle*, blue, mark options={scale=1},
  error bars/.cd, 
    y fixed,
    y dir=both, 
    y explicit
] table [x=x, y=y,y error=error, col sep=comma] {
    x,    y,       error
    2, 0.9107, 0.0210
    5, 0.5611, 0.0606
    10, 0.4829, 0.0741
    15, 0.5689, 0.1562
    20, 0.0008, 0.0006
    25, 0.0007, 0.0004

};
\addlegendentry{\words ``thirty-two''}

\addplot[
  black, mark=*, gray, mark options={scale=1},
  error bars/.cd, 
    y fixed,
    y dir=both, 
    y explicit
] table [x=x, y=y,y error=error, col sep=comma] {
    x,    y,       error
    2, 0.9985, 0.0025
    5, 0.9377, 0.0186
    10, 0.6178, 0.1230
    15, 0.0089, 0.0247
    20, 0, 0
    25, 0, 0
    30, 0, 0
};
\addlegendentry{\underscore ``3\_2''}

\addplot[
  black, mark=x, green, mark options={scale=1.5},
  error bars/.cd, 
    y fixed,
    y dir=both, 
    y explicit
] table [x=x, y=y,y error=error, col sep=comma] {
    x,    y,       error
    2, 0.6590, 0.0050 
    5, 0.5865, 0.0449
    10, 0.3250, 0.0082
    15, 0.1955, 0.0803
    20, 0, 0
    25, 0, 0
    30, 0, 0
};
\addlegendentry{\fixedCharacter ``0 0 3 2''}

\addplot[
  black, mark=square*, orange, mark options={scale=1},
  error bars/.cd, 
    y fixed,
    y dir=both, 
    y explicit
] table [x=x, y=y,y error=error, col sep=comma] {
    x,    y,       error
    2, 0.9978, 0.002 
    5, 0.9027, 0.019
    10, 0.2792, 0.3042
    15, 0, 0
    20, 0, 0
    25, 0, 0
    30, 0, 0
};
\addlegendentry{\character ``3 2''}

\addplot[
  solid, mark=*, purple, mark options={scale=1},
  error bars/.cd, 
    y fixed,
    y dir=both, 
    y explicit
] table [x=x, y=y,y error=error, col sep=comma] {
    x,    y,       error
    2, 0.3526, 0.02
    5, 0.0003, 0
    10, 0, 0
    15, 0, 0
    20, 0, 0
    25, 0, 0
    30, 0, 0
};
\addlegendentry{\decimal ``32''}

\end{axis}
\end{tikzpicture}
\caption{Accuracy of different number representations on the addition task.}
\label{figure:orthography}
\end{figure*}
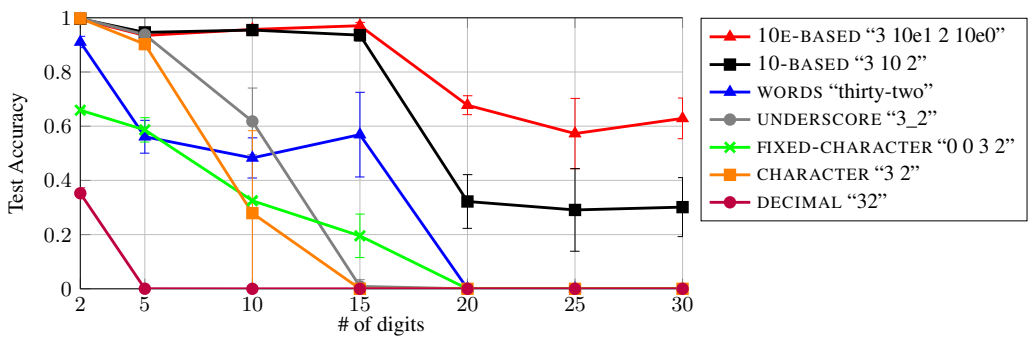

We present results in Figure~\ref{figure:orthography}.
Each point in the graph represents the mean accuracy of a T5-220M model trained for 100 epochs with five different sets of 1,000 addition examples sampled using the balanced method.
A separate development set of 1,000 examples is used to select the best checkpoint of each run.
Error bars correspond to 95\% confidence intervals.
The values on the $x$-axis represent the maximum number of digits used for training and testing.
We use a maximum of 30-digit numbers as some representations such as \words would result in input sequences that have too many tokens (e.g., more than 512), and hence prohibitively long training times.

In the \decimal representation, the model barely learns addition of 2-digit numbers, and it fails to learn addition of larger numbers, i.e., it has an accuracy of zero for 5 digits or more.
One explanation for this failure is because numbers are not systematically tokenized into digits.
For instance, ``132'' might be tokenized as ``1'' and ``32'', whereas ``232'' might be tokenized as ``23'' and ``2''.
Hence, the model would have to learn that sometimes the embedding of a token refers to a single digit, other times to two digits, etc.
It might be hard to learn (i.e., need more examples) to map an embedding to a number when the number of digits it represents changes irregularly (dependent on the training data of the tokenizer).

The \character and \underscore representations have much higher accuracy than \decimal, thus showing that it is easier to learn when embeddings represent single digits.
Both representations exhibit decreasing accuracy as we increase the number of digits, until reaching an accuracy of zero with 15-digit addition.
One explanation for this failure is that, since digits with the same significance have different positions in each example, the model has to count the number of digits on the right side in order to find its significance.
With larger numbers, counting becomes harder.

The \fixedCharacter representation achieves higher accuracy than \character and \underscore for numbers longer than 12 digits, thus showing that the model can learn to memorize digit positions to determine their significance.
However, with an accuracy of approximately 20\% for 15-digit numbers, the memorization strategy eventually breaks down.
It appears to be hard to learn relative positional embeddings that precisely encode the distance between two tokens for our task. 

The \words representation shows stable accuracy in the range of 40-60\% from 5 to 15 digits.
Our hypothesis for this stability is that the intrinsic position tokens present in this representation (e.g., ``hundred'', ``thousand'') make it easier for the model to find and sum two digits that are far apart in the input sequence.
However, for 20 digits or more, the models fail at the task.
Pretraining might have contributed to the high accuracy on 15 digits or less because the model might have already seen these numbers in this representation in the pretraining corpus.
On the other hand, it is very unlikely that the corpus contains numbers of 20 digits or more expressed in plain English.
We further investigate the impact of pretraining in Appendix~\ref{section:pretrained_scratch}.

With up to 15 digits, the \tenBased and \tenEBased representations achieve accuracy close to 100\%.
Our explanation for their success is the explicit position tokens added between each digit, which allows the model to inspect the left or right tokens of a digit to determine its significance.

In the Appendices, we present a number of additional experimental results that build on our main findings here.
In Appendix~\ref{section:position_embeddings}, we study the impact of various position embeddings on the addition task.
In Appendix~\ref{section:interpolation_extrapolation}, we investigate how models of different sizes perform interpolation and extrapolation tasks. Although larger models perform better than smaller ones, we show that not even 3B-parameter models can learn simple arithmetic rules.
In Appendix~\ref{section:data_size_impact}, we show that all representations can reach accuracies of 97\% or more when enough training data is provided. Results here, however, show that representations do matter when training data is scarce.
In Appendices~\ref{section:pretrained_scratch} and~\ref{section:bases}, we study how pretraining can impact a model's ability to learn arithmetic.
Finally, in Appendix~\ref{section:out_of_distribution}, we investigate how a mismatch between the length distribution of training and test sets can be problematic for the addition task.

\section{Conclusion}

\citet{rumelhart1985learning} wrote in their germinal ``backpropagation'' paper that ``unfortunately, this [addition] is the one problem we have found that reliably leads the system into local minima''.
Almost four decades later, despite remarkable progress in neural networks, the field is still exploring this task.
Our small contribution is to show that simple manipulations of surface representations to render semantics explicit can help neural models to learn simple arithmetic tasks. 
It remains to be seen if this ``trick'' can be applied to other tasks, but our results provide evidence that improving tokenizers and positional encodings are promising directions for future exploration.

\section*{Acknowledgments}

This research was supported in part by the Canada First Research Excellence Fund and the Natural Sciences and Engineering Research Council (NSERC) of Canada.
In addition, we would like to thank Google Cloud for credits to support this work.

\bibliography{main}
\bibliographystyle{iclr2021_conference}

\clearpage
\appendix

\section{Related Work}
\label{section:related_work}

Recent studies have explored the numerical capabilities learned by neural networks trained on large amounts of texts~\citep{talmor2019olmpics,jiang2019learning,naik2019exploring,wallace2019nlp,lin2020birds,johnson2020probing,mishra2020towards}.
See~\citet{thawani2021representing} for a detailed survey.

A common finding is that the learned embeddings capture \textit{magnitude} (e.g., 2 < 3), but many models fail to capture \textit{numeracy} (e.g., two=2)~\citep{naik2019exploring,wallace2019nlp,ren2020enhancing,zhang2020language}.
Character-level models such as ELMO~\citep{peters2018deep} have stronger numeracy than sub-word models such as BERT~\citep{devlin2019bert}, perhaps because two numbers that are similar in value can have very different sub-word tokenizations~\citep{wallace2019nlp}.
Our work shows that characters are adequate representations for small to medium numbers, but they are not sufficient when dealing with large numbers, which require precise position representations for each digit.

However, independently of the tokenization method, pretrained word embeddings have trouble extrapolating to numbers unseen during training~\citep{wallace2019nlp}.
Some alternatives to improve the extrapolation capabilities of neural models include augmenting pretraining corpora with numerical texts~\citep{geva-etal-2020-injecting,chu2020learning} or using scientific notation to represent numbers~\citep{zhang2020language}.
Similarly, better numerical skills can be achieved by augmenting input texts with pre-computed numerical computations~\citep{andor2019giving} or by explicitly inferring mathematical equations from natural language text~\citep{zou2019quantity,zou2019text2math,li2019modeling,liu2019tree,shi2020sequence}.

Special architectures have also been proposed for arithmetic tasks~\citep{kaiser2015neural,kalchbrenner2015grid,price2016extensions,trask2018neural}.
Many of these models are capable of summing numbers larger than the ones seen during training.
In contrast, more general-purpose architectures fail to extrapolate on numerical tasks~\citep{joulin2015inferring,dehghani2018universal,schlag2019enhancing}.

Others have proposed neural--symbolic hybrids, which are typically composed of a neural model to convert inputs to contiguous vector representations and a symbolic component that applies rules over these vectors~\citep{ran2019numnet}.
However, a body of evidence has shown that neural networks can perform reasoning tasks.
For instance, a modern pretrained model with self-attention that uses the right level of input representation can outperform neural--symbolic hybrids on artificial reasoning tasks that require answering questions from videos~\citep{ding2020object}.
Deep learning models were also successfully applied to symbolic integration, to solve differential equations~\citep{lample2019deep}, and automated theorem proving~\citep{polu2020generative}.

Furthermore, it is not clear how architectures specialized to some tasks can be adapted to simultaneously perform a range of tasks a human is capable of.
Our work instead focuses on a general-purpose architecture that can be applied to almost all natural language processing tasks.

Novel ways of encoding positions of tokens in the transformer architecture have been proposed, but they were mostly evaluated on natural language processing tasks, showing small performance gains~\citep{ke2020rethinking,he2020deberta,wang2019encoding,huang2020improve}.
We instead expose the limitations of subword tokenizers and positional encodings using simple arithmetic tasks.

Datasets such as DROP~\citep{Dua2019DROPAR}, EQUATE~\citep{ravichander2019equate}, or Mathematics Questions~\citep{saxton2018analysing} test numerical reasoning; they contain examples that require comparing, sorting, and performing other complex mathematical tasks.
This work focuses on isolating the failure cases of the transformer architecture by studying how it performs simple arithmetic tasks.
We argue that this is a necessary skill to solve more complex reasoning tasks.

\section{Position Embeddings}
\label{section:position_embeddings}

Here, we study the impact of various position embeddings on the addition task.
Since pretraining from scratch is a costly process, we experiment with only small transformer models fine-tuned without pretraining.

The architecture of the transformer follows \citet{vaswani2017attention} except we use 4 layers for the encoder and the decoder, respectively.
We look into the effect of representation and positional encoding on addition from 2 digits to 9 digits.
Due to the cost of these experiments, we choose a subset of the representations studied in Section~\ref{section:orthography_results}:\ \tenEBased, \tenBased, and \character.

The dataset is split into training and test sets with a ratio of 9:1. 
For 3--9 digits addition, we randomly generate 10,000 samples for the whole dataset.
For 2-digit addition, we use all of the combinations for every addend $a\in [10, 99]$, which results in less than 10,000 samples.
The models are trained for 55 epochs with a learning rate of $10^{-5}$. 

\begin{figure*}[t]
\centering
\begin{tikzpicture}[scale = 0.88]
\begin{axis}[
width=0.8\textwidth,
height=0.30\textwidth,
legend cell align=left,
mark options={mark size=3},
font=\scriptsize,
axis y line*=left,
xmin=2, xmax=9,domain=1:10,
ymin=0.0, ymax=1.0,
every axis plot/.append style={thick},
xtick={2,3,4,5,6,7,8,9},
ytick={0.0, 0.2, 0.4, 0.6, 0.8, 1.0},
legend pos=outer north east,
xmajorgrids=true,
ymajorgrids=true,
xlabel style={font = \small, yshift=1ex},
xlabel=\# of digits,
ylabel= Test Accuracy,
ylabel style={font = \small, yshift=0ex}]

\addplot[
  black, mark=*, red, mark options={scale=1},
  error bars/.cd, 
    y fixed,
    y dir=both, 
    y explicit
] table [x=x, y=y, col sep=comma] {
    x,    y
    2, 1.000
    3, 0.983
    4, 0.768
    5, 0.622
    6, 0.814
    7, 0.379
    8, 0.149
    9, 0.006
};
\addlegendentry{\textsc{10e, Pos-Masked, With Tgt}}

\addplot[
  black, mark=*, blue, mark options={scale=1},
  error bars/.cd, 
    y fixed,
    y dir=both, 
    y explicit
] table [x=x, y=y, col sep=comma] {
    x,    y
    2, 0.983
    3, 0.663
    4, 0.385
    5, 0.190
    6, 0.025
    7, 0.005
    8, 0.000
    9, 0.000
};
\addlegendentry{\textsc{10, Pos-Masked, With Tgt}}

\addplot[
  black, mark=*, black, mark options={scale=1},
  error bars/.cd, 
    y fixed,
    y dir=both, 
    y explicit
] table [x=x, y=y, col sep=comma] {
    x,    y
    2, 0.538
    3, 0.048
    4, 0.002
    5, 0.000
    6, 0.000
    7, 0.000
    8, 0.000
    9, 0.000
};
\addlegendentry{\textsc{char, Pos-Masked, With Tgt}}

\addplot[
  black, mark=square*, green, mark options={scale=1},
  error bars/.cd, 
    y fixed,
    y dir=both, 
    y explicit
] table [x=x, y=y, col sep=comma] {
    x,    y
    2, 1.000
    3, 0.985
    4, 0.971
    5, 0.91
    6, 0.884
    7, 0.824
    8, 0.726
    9, 0.554
};
\addlegendentry{\textsc{10e, Pos-Masked, No Tgt}}

\addplot[
  black, mark=square*, gray, mark options={scale=1},
  error bars/.cd, 
    y fixed,
    y dir=both, 
    y explicit
] table [x=x, y=y, col sep=comma] {
    x,    y
    2, 0.996
    3, 0.988
    4, 0.941
    5, 0.789
    6, 0.742
    7, 0.384
    8, 0.000
    9, 0.000
};
\addlegendentry{\textsc{10, Pos-Masked, No Tgt}}

\addplot[
  black, mark=square*, orange, mark options={scale=1},
  error bars/.cd, 
  y fixed,
  y dir=both,
  y explicit
] table [x=x, y=y, col sep=comma] {
    x,    y
    2, 1.000
    3, 0.996
    4, 0.971
    5, 0.913
    6, 0.825
    7, 0.601
    8, 0.452
    9, 0.017
};
\addlegendentry{\textsc{char, Pos-Masked, No Tgt}}

\addplot[
  black, dashed, mark=triangle*, green, mark options={scale=1},
  error bars/.cd, 
    y fixed,
    y dir=both, 
    y explicit
] table [x=x, y=y, col sep=comma] {
    x,    y
    2, 0.242
    3, 0.021
    4, 0.000
    5, 0.000
    6, 0.000
    7, 0.000
    8, 0.000
    9, 0.000
};
\addlegendentry{\textsc{10e, Sinusoidal}}

\addplot[
  black, dashed, mark=triangle*, gray, mark options={scale=1},
  error bars/.cd, 
    y fixed,
    y dir=both, 
    y explicit
] table [x=x, y=y, col sep=comma] {
    x,    y
    2, 0.252
    3, 0.022
    4, 0.000
    5, 0.000
    6, 0.000
    7, 0.000
    8, 0.000
    9, 0.000
};
\addlegendentry{\textsc{10, Sinusoidal}}

\addplot[
  black, dashed, mark=triangle*, orange, mark options={scale=1},
  error bars/.cd, 
    y fixed,
    y dir=both, 
    y explicit
] table [x=x, y=y,col sep=comma] {
    x,    y
    2, 0.219
    3, 0.019
    4, 0.000
    5, 0.000
    6, 0.000
    7, 0.000
    8, 0.000
    9, 0.000
};
\addlegendentry{\textsc{char, Sinusoidal}}

\end{axis}
\end{tikzpicture}
\caption{Addition accuracy of vanilla transformers with different position encoding methods.}
\label{figure:various_position}
\end{figure*}
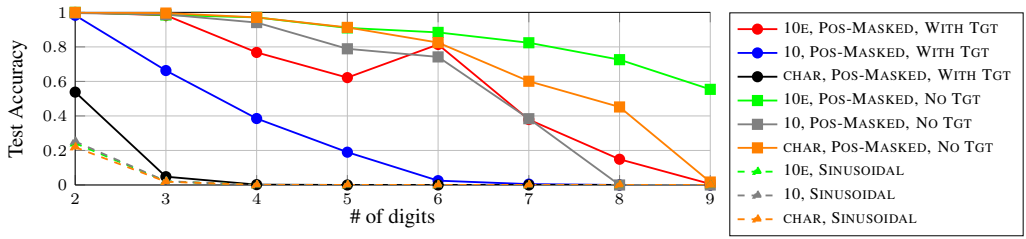

We find that the original positional encoding in~\citet{vaswani2017attention} fails to learn addition effectively, as shown in Figure~\ref{figure:various_position}.
This might be due to the correlation introduced by two heterogeneous signals---embedding and absolute positional encoding~\citep{ke2020rethinking}.
Therefore, we designed a position-wise masked embedding for this task.

More specifically, for an $n$-digit number whose embedding is $e$ with embedding size $d$, we will set $e[u:v]=1$ for $i-th$ digit in the number, where $u=int(\frac{d}{n}) \cdot (n-i)$ and $v=int(\frac{d}{n}) \cdot (n-i+1)$.
We set other position embedding values to 0.
Note that $i$ follows the ``Big-Endian'' style (e.g., $i=3$ for ``2'' in the number ``271'').
However, during inference, digit information is not provided for the target sequence as we don't know the exact digit of the decoded number in advance.
So, we face a format discrepancy between training and inference.
To investigate how this discrepancy will affect the result, we train the model in two different ways---training with target position provided and training without target position provided (position encoding for the target is the zero vector).
Note that position encoding is provided for the source sequence in both cases for training and inference; position encoding is not provided for the target sequence during inference in both cases.
The results are shown in Figure~\ref{figure:various_position}, labeled as ``WITH TGT'' and ``NO TGT'', respectively. We label our position-wise masked embedding as ``Pos-Masked''. The original representation is called ``Sinusoidal''.

Consistent with previous experiments, \tenEBased performs best given the same position encoding and training strategies.
Comparing ``WITH TGT'' and ``NO TGT'', we can see that training with target position encoding creates fluctuations among different digits.
In general, it performs worse than training without target position encoding given the same encoding representation.
Unsurprisingly, under our experiment setting, whether the target position is provided is not as important as having the same format between training and inference.

\section{Experiments on Extrapolation}
\label{section:interpolation_extrapolation}

One advantage of working with arithmetic tasks is that the rules to be learned are well defined and relatively simple.
Thus, it is easy to verify if models learned such rules by evaluating them on numbers that are larger than the ones they were trained on.
If successful, such a model would have no problem correctly adding or subtracting arbitrarily long numbers. 

In this section, we investigate how models of different sizes perform interpolation and extrapolation tasks.
We train T5-60M, T5-220M, T5-770M, and T5-3B models on numbers that are sampled using the ``balanced'' method.
Models are trained 100K iterations using batches of 128 examples and a learning rate of $10^{-3}$.
We save checkpoints every 2,000 iterations, and the best checkpoint is chosen using a separate validation set of 10,000 examples.
The models are evaluated on a test set of 10,000 examples with numbers sampled using the ``random'' method.

For interpolation experiments, the models are trained and evaluated on up to 60-digit numbers. 
For extrapolation experiments, the models are trained on up to 50-digit numbers and evaluated on 60-digit numbers.
We use that many digits for training because the models could not extrapolate with fewer; see more below.

{\bf Regular vs.\ inverse orders:} Auto-regressive models such as the ones used in this work generate the output sequence token by token.
Thus, to produce the first digit of the answer, which is the most significant one, the model has to perform all the carry operations.
In the addition example ``What is 52 plus 148?'', to produce the first digit ``2'', the model has to perform the carry operation for the unit digits (2 and 8), and then the carry for the decimal digits (5 and 4).
Hence, the model has to perform the digit-wise addition (or subtraction) of all the digits in the question before generating the first digit of the answer.
We call this generation order ``regular''.

Another way to produce an answer is by generating the least significant digits first.
This order is perhaps easier to learn than the ``regular'' order because to decode each digit, the model only needs to add (or subtract) single digits and check if the previous digit-wise operation had a carry.
We call this generation order ``inverse''.

The results presented in Table~\ref{table:interpolation_extrapolation} show that models of all sizes successfully perform interpolation tasks.
Two exceptions are T5-60M on the subtraction tasks, which achieve 0.934 and 0.830 accuracy for inverse and regular orders, respectively.
Nevertheless, compared to the extrapolation results, these numbers are high enough to consider them as successful runs.

On extrapolation tasks, T5-3B succeeds on almost all of them, whereas smaller models fail more often.
Even on tasks where T5-220M achieves reasonable accuracy (0.862 and 0.641 on addition and subtraction using regular order, respectively), T5-3B outperforms T5-220M by large margins.
This result provides evidence that larger models might perform better on data whose distribution is outside its training data distribution.
However, it remains to be investigated if this trend holds for more complex tasks, especially those involving natural language.

The difference in accuracy is negligible between regular and inverse orders on interpolation tasks.
However, models trained and evaluated on the regular order show higher extrapolation accuracy than those that use the inverse order.
For example, T5-220M fails to extrapolate on both addition and subtraction tasks when using the inverse order (i.e., accuracy is zero), but it performs better when using the regular order, with accuracy between 60--90\%.
This result is perhaps surprising since one would expect that the inverse order would be easier to learn.

Supported by recent work, we suspect that the problem is related to the bias of selecting the termination (i.e., end-of-sequence) token when the generated sequence becomes longer than those seen during training~\citep{newman2020eos}.
In the inverse order, the answer is generated from least to most significant digit, so the model might have a tendency to select the termination token right after it generates the most significant digit seen during training.
In the regular order, however, the model has to predict the full length of the sequence before emitting the first and second tokens.
For example, the first two tokens of the answer to the question $10^{60} + 10^{60}$ are ``2'' and ``10e60''.
This explicit length prediction allows the model to better generalize to longer sequences, but it appears to be insufficient to induce models to learn addition rules that are independent of the length of numbers seen during training (more below).

\begin{table*}[t]
\centering\centering
\begin{tabular}{lcccc|cccc}
\midrule
& \multicolumn{4}{c|}{Interpolation} & \multicolumn{4}{c}{Extrapolation}\\
Order: & \multicolumn{2}{c}{Inverse} & \multicolumn{2}{c|}{Regular} & \multicolumn{2}{c}{Inverse} & \multicolumn{2}{c}{Regular}\\
Operation: & Add & Sub & Add & Sub & Add & Sub & Add & Sub \\
\toprule
T5-60M & \textbf{1.000} & 0.934 & \textbf{0.998} & 0.830 & 0.000 & 0.000 & 0.004 & 0.000 \\
T5-220M & \textbf{1.000} & \textbf{0.998} & \textbf{1.000} & \textbf{0.995} & 0.000 & 0.000 & 0.862 & 0.641 \\
T5-770M & \textbf{1.000} & 0.947 & \textbf{0.999} & \textbf{0.982} & 0.003 & 0.000 & 0.442 & 0.373 \\
T5-3B & \textbf{1.000} & \textbf{0.997} & \textbf{1.000} & \textbf{0.993} & \textbf{0.974} & 0.865 & \textbf{0.988} & \textbf{0.982} \\
\bottomrule
\end{tabular}
\vspace{0.15cm}
\caption{Interpolation and extrapolation accuracy. Interpolation refers to training and testing on up to 60-digit numbers. Extrapolation refers to training on up to 50-digit numbers and testing on 60-digit numbers. We highlight in \textbf{bold} accuracy above 97\%.}
\label{table:interpolation_extrapolation}
\vspace{0.0cm}
\end{table*}


We observe high variance in accuracy for the extrapolation experiments.
For example, during the training of a T5-770M model on up to 30-digit numbers, the accuracy ranges from 20\% to 50\% when evaluated on 60-digit numbers.
Extrapolation accuracy also oscillates between 20--40 percentage points when changing the seed for training data generation.

Extrapolation is hardly achieved when trained on fewer than 50 digits, regardless of the model size.
For example, T5-220M, T5-770M, and T5-3B trained on 15 digits show an accuracy of zero when evaluated on 20 digits.

Beyond a critical amount, increasing the training data does not improve extrapolation accuracy.
For example, when trained on up to 30-digit and evaluated on 60-digit numbers, a T5-770M showed a similar accuracy range (20\%--50\%) when trained with either 100K, 1M, or 10M examples.
As training progresses, interpolation accuracy always reaches 100\%, but extrapolation accuracy starts to decrease after some number of training steps.
The number of training steps after which this drop occurs varies dramatically between runs that differ only in the seed used to generate the training data.
We are unable to isolate the cause of this behavior.

Contrary to the hypothesis of \citet{newman2020eos}, we find that the end-of-sequence token does not seem to be the cause of extrapolation failures.
For example, when a T5-770M model trained on 30-digit numbers is evaluated on 60-digit numbers, it correctly generates the first 23 position tokens (i.e., from ``10e60'' until ``10e38'') but it suddenly skips to position token ``10e27'', and continues generating the correct position tokens until the last one (``10e0''). Here we show one such sequence:

\begin{quote}
1 10e60 0 10e59 1 10e58 2 10e57 3 10e56 0 10e55 2 10e54 7 10e53 0 10e52 1 10e51 0 10e50 3 10e49 9 10e48 0 10e47 5 10e46 3 10e45 1 10e44 5 10e43 3 10e42 6 10e41 3 10e40 6 10e39 0 \textbf{\red{10e38}} 8 \textbf{\red{10e27}} 1 10e26 4 10e25 1 10e24 2 10e23 6 10e22 6 10e21 9 10e20 5 10e19 3 10e18 4 10e17 8 10e16 3 10e15 8 10e14 8 10e13 9 10e12 5 10e11 3 10e10 5 10e9 0 10e8 6 10e7 4 10e6 3 10e5 5 10e4 6 10e3 7 10e2 2 10e1 2 10e0
\end{quote}

Hence, although the model correctly emits the end-of-sequence token after the ``10e0'' token, it decides to shorten the sequence in the middle of the generation, i.e., by skipping position tokens ``10e37'' until ``10e28''.
This skipping behavior is consistent across model sizes, dataset sizes, and extrapolation ranges (e.g., training on 20 digits, evaluating on 30 digits, etc.).
Investigating it further might help us understand why neural models often fail on extrapolation tasks.

\section{Impact of data size}
\label{section:data_size_impact}

In Section~\ref{section:orthography_results}, we show that the choice of orthography has a large impact on the addition task when training data is scarce (i.e., 1,000 training examples).
In this section, we investigate how these representations perform with varying amounts of training data.
We train and evaluate T5-220M on the addition task of up to 30-digit numbers using the regular order.
Due to the high computational cost of training this model on millions of examples, we reduce the number of epochs depending on the dataset size, which is detailed in Table~\ref{table:size_epochs}.
We select the best checkpoint using a validation set of 10,000 examples and evaluate the models on a test set of 10,000 examples.

\begin{table}[h]
\centering\centering
\begin{tabular}{lr}
\midrule
Size & Epochs\\
\toprule
$10^3$ & 200 \\
$10^4$ & 100 \\
$10^5$ & 20 \\
$10^6$ & 10 \\
$10^7$ & 1 \\
\bottomrule
\end{tabular}
\vspace{0.15cm}
\caption{Number of training epochs for each dataset size presented in Figure~\ref{figure:training_size}.}
\label{table:size_epochs}
\end{table}

\begin{figure*}[h!]
\centering
\begin{tikzpicture}[scale = 0.8]
\begin{axis}[
width=0.90\textwidth,
height=0.40\textwidth,
legend cell align=left,
font=\small,
axis y line*=left,
xmode=log,
log ticks with fixed point,
every axis plot/.append style={very thick},
xmin=1000, xmax=10000000,domain=1:10,
ymin=0.0, ymax=1.0,
xtick={1000, 10000, 100000, 1000000, 10000000},
ytick={0.0, 0.2, 0.4, 0.6, 0.8, 1.0},
legend pos=south east,
xmajorgrids=true,
ymajorgrids=true,
xlabel style={font = \small, yshift=1ex},
xlabel=\# Training examples,
ylabel= Test Accuracy,
ylabel style={font = \small, yshift=0ex}]

\addplot[
  black, mark=triangle*, red, mark options={scale=1},
  error bars/.cd, 
    y fixed,
    y dir=both, 
    y explicit
] table [x=x, y=y,y error=error, col sep=comma] {
    x,    y,       error
    1000, 0.6288, 0.0390
    10000, 0.9991, 0
    100000, 0.9995, 0
    1000000, 0.9999, 0
    10000000, 0.9999, 0
};
\addlegendentry{\tenEBased ``3 10e1 2 10e0''}

\addplot[
  black, mark=square*, black, mark options={scale=1},
  error bars/.cd, 
    y fixed,
    y dir=both, 
    y explicit
] table [x=x, y=y,y error=error, col sep=comma] {
    x,    y,       error
    1000, 0.3014, 0.1088
    10000, 0.9839, 0
    100000, 0.9976, 0
    1000000, 0.9995, 0
    10000000, 1.0000, 0
};
\addlegendentry{\tenBased ``3 10 2''}

\addplot[
  solid, mark=triangle*, blue, mark options={scale=1},
  error bars/.cd, 
    y fixed,
    y dir=both, 
    y explicit
] table [x=x, y=y,y error=error, col sep=comma] {
    x,    y,       error
    1000, 0.0005, 0
    10000, 0.9692, 0
    100000, 0.9794, 0
    1000000, 0.9999, 0
    10000000, 0.9999, 0
};
\addlegendentry{\words ``thirty-two''}

\addplot[
  black, mark=*, gray, mark options={scale=1},
  error bars/.cd, 
    y fixed,
    y dir=both, 
    y explicit
] table [x=x, y=y,y error=error, col sep=comma] {
    x,    y,       error
    1000, 0, 0
    10000, 0.7977, 0
    100000, 0.9963, 0
    1000000, 0.9972, 0
    10000000, 0.9998, 0
};
\addlegendentry{\underscore ``3\_2''}


\addplot[
  black, mark=square*, orange, mark options={scale=1},
  error bars/.cd, 
    y fixed,
    y dir=both, 
    y explicit
] table [x=x, y=y,y error=error, col sep=comma] {
    x,    y,       error
    1000, 0, 0
    10000, 0.7339, 0
    100000, 0.991, 0
    1000000, 0.9979, 0
    10000000, 0.9998, 0
};
\addlegendentry{\character ``3 2''}

\addplot[
  solid, mark=*, purple, mark options={scale=1},
  error bars/.cd, 
    y fixed,
    y dir=both, 
    y explicit
] table [x=x, y=y,y error=error, col sep=comma] {
    x,    y,       error
    1000, 0, 0
    10000, 0, 0
    100000, 0.8855, 0
    1000000, 0.9653, 0
    10000000, 0.9788, 0
};
\addlegendentry{\decimal ``32''}

\end{axis}
\end{tikzpicture}
\caption{Accuracy of different number representations when varying the amount of training examples. The task is addition of 30-digit numbers.}
\label{figure:training_size}
\end{figure*}
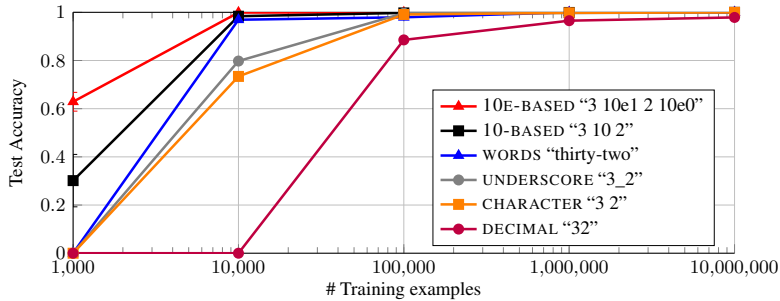

Results are shown in Figure~\ref{figure:training_size}.
The \tenEBased representation presents the best results for training sizes of 1,000 and 10,000 examples, followed by \tenBased, \words, \underscore, \character, and \decimal.
For larger datasets such as 10M examples, almost all representations achieve more than 99.9\% accuracy.
The exception is the \decimal representation, which still has a high error of 2.1\% even when trained with 10M examples.

We conclude that with enough training data, models can learn the addition task regardless of the representation.
The limitations of some representations are exposed only when training data is small.

\section{Pretrained vs.\ From Scratch Models}
\label{section:pretrained_scratch}

One hypothesis for the high interpolation accuracy reported in Section~\ref{section:orthography_results} despite using a small number of training examples is that the model has already seen addition and subtraction examples during pretraining.
To test this hypothesis, we compare pretrained models with models trained from scratch (i.e., no pretraining on the masked language modeling task) on the addition task.
In this experiment, the models never see the same training example more than once.
That is, they are not limited by training data. 

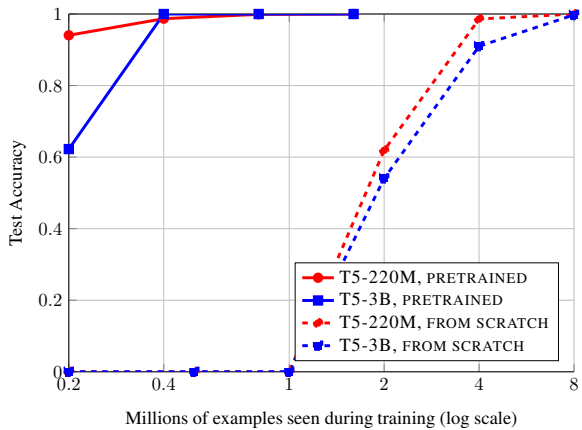
\begin{figure}[h]
\centering
\begin{tikzpicture}[scale = 0.7]
\begin{axis}[
width=0.8\textwidth,
height=0.60\textwidth,
legend cell align=left,
mark options={mark size=3},
axis y line*=left,
xmode=log,
log ticks with fixed point,
every axis plot/.append style={ultra thick},
xmin=0.2, xmax=8,domain=1:10,
ymin=0.0, ymax=1.0,
xtick={0.2, 0.4, 1, 2, 4, 8},
ytick={0.0, 0.2, 0.4, 0.6, 0.8, 1.0},
legend pos=south east,
xmajorgrids=true,
ymajorgrids=true,
xlabel style={yshift=-1ex},
xlabel=Millions of examples seen during training (log scale),
ylabel= Test Accuracy,
ylabel style={ yshift=0ex}]

\addplot[
  black, mark=*, red, mark options={scale=1},
  error bars/.cd, 
    y fixed,
    y dir=both, 
    y explicit
] table [x=x, y=y,y error=error, col sep=comma] {
    x,    y,       error
    0.2, 0.9405, 0
    0.4, 0.987, 0
    0.8, 0.999, 0
    1.6, 1, 0 
};
\addlegendentry{\textsc{T5-220M, pretrained}}

\addplot[
  black, mark=square*, blue, mark options={scale=1},
  error bars/.cd, 
    y fixed,
    y dir=both, 
    y explicit
] table [x=x, y=y,y error=error, col sep=comma] {
    x,    y,       error
    0.2, 0.6225, 0
    0.4, 1, 0
    0.8, 1, 0
    1.6, 1, 0 
};
\addlegendentry{\textsc{T5-3B, pretrained}}

\addplot[
  black, dashed, mark=*, red, mark options={scale=1},
  error bars/.cd, 
    y fixed,
    y dir=both, 
    y explicit
] table [x=x, y=y,y error=error, col sep=comma] {
    x,    y,       error
    0.2, 0, 0
    0.5, 0, 0
    1.0, 0, 0
    2.0, 0.619, 0
    4.0, 0.986, 0
    8.0, 1, 0
};
\addlegendentry{\textsc{T5-220M, from scratch}}

\addplot[
  black, dashed, mark=square*, blue, mark options={scale=1},
  error bars/.cd, 
    y fixed,
    y dir=both, 
    y explicit
] table [x=x, y=y,y error=error, col sep=comma] {
    x,    y,       error
    0.2, 0, 0
    0.5, 0, 0
    1.0, 0, 0
    2.0, 0.5405, 0
    4.0, 0.91, 0
    8.0, 0.9975, 0
    12.0, 1, 0
};
\addlegendentry{\textsc{T5-3B, from scratch}}

\end{axis}
\end{tikzpicture}
\caption{Accuracy of pretrained models vs.\ from scratch models with respect to the number of training examples. Models are trained and evaluated on numbers with up to 60 digits in length.}
\label{figure:pretrained_scratch}
\end{figure}

Figure~\ref{figure:pretrained_scratch} shows that both pretrained T5-220M and T5-3B need approximately ten times fewer training examples (and compute) than models trained from scratch to reach 100\% accuracy on the addition of 60-digit numbers.

\section{Accuracy on Different Bases}
\label{section:bases}

Here we propose another way to test how pretraining can impact a model's ability to learn arithmetic. 
We hypothesize that a model might have difficulty learning bases different than base 10 (i.e., decimal) because examples rarely occur in the pretraining corpus.
To test this hypothesis, we train a T5-220M model on addition examples using binary, ternary, decimal, and base 19.
While there might be examples of binary addition in the pretraining corpus, our expectation is that it contains few (if any?) examples of addition using base 19 numbers.
We use the \tenEBased orthography and inverse order due to its slightly better accuracy (see Table~\ref{table:interpolation_extrapolation}).
We also evaluate models trained from scratch.

We report the mean accuracy and 95\% confidence intervals of a model trained with five different sets of 1,000 addition examples for 100 epochs.
A separate development set of 1,000 examples was used to select the best checkpoint of each run.
We trained and evaluated on numbers equivalent to 15 decimal digits.

For these experiments, we use only 1,000 training examples since experiments in Appendix~\ref{section:data_size_impact} show that models can successfully learn with enough training data, thus too much data defeats the purpose of measuring the impact of pretraining; see also \citet{hernandez2021scaling}.
Results are shown in Table~\ref{table:various_bases}.
The pretrained model has no problem learning binary, ternary, and decimal bases, but its accuracy degrades slightly on base 19.
Since it is unlikely that the pretrained model has encountered substantial numbers of examples of addition in rare bases (i.e., ternary and 19), it seems that pretraining helps on this task in other ways than simple memorization.

To show that the task is not easy, we also report in the table that models trained from scratch fail to learn the task regardless of the base.
This result is expected since a large number of parameters (220M) need to be learned from scratch using just 1,000 examples.

\begin{table}
\centering
\begin{tabular}{lcc}
\midrule
 & \multicolumn{2}{c}{Test Accuracy}\\
Base & From Scratch & Pretrained  \\
\toprule
2 & 0.000 $\pm$ 0.000 & \textbf{0.999} $\pm$ 0.001 \\
3 & 0.000 $\pm$ 0.000 & \textbf{0.999} $\pm$ 0.002 \\
10 & 0.000 $\pm$ 0.000 & \textbf{0.993} $\pm$ 0.003 \\
19 & 0.000 $\pm$ 0.000 & \textbf{0.976} $\pm$ 0.007 \\
\bottomrule
\end{tabular}
\vspace{0.15cm}
\caption{Test set accuracy of 15-digit addition on various bases. Numbers are represented with \tenEBased orthography.}
\label{table:various_bases}
\vspace{-0.5cm}
\end{table}

\section{Impact of Different Length Distributions}
\label{section:out_of_distribution}

Here we investigate to what extent a mismatch between the length distribution of training and test sets is problematic for the addition task.
We train T5-220M models on 100,000 examples, select the best checkpoint using a development set of 10,000 examples, and evaluate on another 10,000 examples.
Here we use the regular order.
Training and test sets are generated using either the balanced or random sampling methods described in Section~\ref{section:methodology}.

Results are shown in Table~\ref{table:out_distribution}.
When trained on the balanced distribution, the model succeeds on both random and balanced evaluation sets.
When trained on the random distribution, it succeeds on the random evaluation set, but it fails on the balanced evaluation set.
In other words, when trained on data where most numbers (i.e., 90\%) have 60 digits, it does not learn to add numbers with fewer digits.
This shows that models have problems performing addition of sequences shorter than the ones seen during training.
This is complementary to the results presented in Appendix~\ref{section:interpolation_extrapolation}, which shows that models cannot generate examples longer than the ones seen during training.

\begin{table}[h]
\centering
\begin{tabular}{llcc}
\midrule
& & \multicolumn{2}{c}{Test} \\
& & Balanced & Random\\
\toprule
\multirow{2}{*}{Train} & Balanced & 1.000 & 1.000 \\
 & Random & 0.014 & 1.000 \\
\bottomrule
\end{tabular}
\vspace{0.15cm}
\caption{Accuracy on 60-digit addition, with balanced and random sampling as described in Section~\ref{section:methodology}.}
\label{table:out_distribution}
\vspace{0.0cm}
\end{table}

\end{document}